\documentclass[lettersize,journal]{IEEEtran}
\usepackage{amsmath,amsfonts}
\usepackage{algorithmic}
\usepackage{algorithm}

\usepackage{array}
\usepackage[caption=false,font=normalsize,labelfont=sf,textfont=sf]{subfig}
\usepackage{textcomp}
\usepackage{stfloats}
\usepackage{url}
\usepackage{verbatim}
\usepackage{graphicx}
\graphicspath{{Fig/}} 
\usepackage{array, booktabs} 
\usepackage{bm}
\usepackage{cite}
\begin{document}

\title{Evolutionary Causal Discovery with Relative Impact Stratification for Interpretable Data Analysis}

\author{%
\IEEEauthorblockN{
Ou Deng\IEEEauthorrefmark{1}\thanks{\IEEEauthorrefmark{1}Graduate School of Human Sciences, Waseda University. },
Shoji Nishimura\IEEEauthorrefmark{2}, Atsushi Ogihara\IEEEauthorrefmark{2}, Qun Jin\IEEEauthorrefmark{2}\thanks{\IEEEauthorrefmark{2}Department of Human Informatics and Cognitive Sciences, Faculty of Human Sciences, Waseda University.}}
}




\maketitle

\begin{abstract}

This study proposes Evolutionary Causal Discovery (ECD) for causal discovery that tailors response variables, predictor variables, and corresponding operators to research datasets. Utilizing genetic programming for variable relationship parsing, the method proceeds with the Relative Impact Stratification (RIS) algorithm to assess the relative impact of predictor variables on the response variable, facilitating expression simplification and enhancing the interpretability of variable relationships.  
ECD proposes an expression tree to visualize the RIS results, offering a differentiated depiction of unknown causal relationships compared to conventional causal discovery. The ECD method represents an evolution and augmentation of existing causal discovery methods, providing an interpretable approach for analyzing variable relationships in complex systems, particularly in healthcare settings with Electronic Health Record (EHR) data.
Experiments on both synthetic and real-world EHR datasets demonstrate the efficacy of ECD in uncovering patterns and mechanisms among variables, maintaining high accuracy and stability across different noise levels. On the real-world EHR dataset, ECD reveals the intricate relationships between the response variable and other predictive variables, aligning with the results of structural equation modeling and shapley additive explanations analyses.

\end{abstract}


\section{Introduction}
\label{sec:Intro}

The systematic analysis of Electronic Health Records (EHRs) has become a cornerstone of contemporary medical and health research, playing a pivotal role in elucidating disease mechanisms, optimizing clinical decision-making processes, and refining policy development \cite{LaCava2023, Kotoku2020, Erion2022, Morin2021}. In the absence of randomized controlled trials, causal discovery and Structural Equation Modeling  (SEM) have emerged as the primary approaches for exploring the intricate relationships within medical data \cite{Hoyle1995, Pearl2000, Glymour2019}. Causal discovery endeavors to deduce potential causal relationships directly from data, typically represented as Directed Acyclic Graphs (DAGs), whereas SEM constructs and validates hypothetical causal models using an array of statistical techniques, including factor, path, and regression analyses \cite{Hoyle1995}.

Despite the theoretical advantages of these methods, their practical implementation is fraught with numerous challenges. First, the requisite sample sizes for the stable operation of causal discovery algorithms often exceed the available volume of clinical data for specific diseases, limiting their applicability in small-sample studies. Second, different causal discovery algorithms may yield divergent causal graphs from the same dataset, including variations in the directionality and strength of causal relationships, and even contradictory causal directions for key feature variables, thus complicating interpretation \cite{Glymour2019, DirectLiNGAM2011}. Third, the effectiveness of SEM considerably depends on the accuracy of model specification and data quality, where inappropriate model assumptions or data issues can lead to misleading conclusions \cite{Hoyle1995}. Addressing these practical challenges necessitates meticulous consideration in the selection and application of these methods, ensuring alignment with research objectives and data characteristics, and the implementation of appropriate measures to mitigate their inherent limitations.

This study proposes an Evolutionary Causal Discovery (ECD) method utilizing Genetic Programming Symbolic Regression (GPSR) to conduct an in-depth analysis of the relationships between predictor and response variables in specific research contexts. The ECD method explores the complex interaction patterns among variables and quantitatively assesses the relative impact effects of predictor variables on response variables, complementing conventional methods like causal discovery and SEM to collectively provide a scientific basis for clinical decision-making processes.

Unlike causal discovery methods that directly seek the causal relationship graphs between variables, the ECD method identifies interaction patterns among variables by unveiling the intrinsic mathematical relationships in data. Compared to SEM, which relies on strict model specification and prior knowledge, the ECD method provides a flexible approach to exploring potential relationships between variables without the need for prior assumptions. Therefore, the ECD method demonstrates unique advantages in handling complex and structurally unknown data, promising to become a novel, interpretable method for complex data analysis, including EHR, and expanding the applicability of causal exploration.

The ECD method proposes an original Relative Impact Stratification (RIS) algorithm to analyze variable relationships by evaluating the impact of minor perturbations in predictor variables across statistical quartiles. The RIS algorithm provides a quantitative basis for expression simplification, refining the articulation of key system mechanisms. ECD utilizes an expression tree, equivalent to parsing expressions, to visualize the RIS results, offering a differentiated depiction of unknown causal relationships compared to conventional causal discovery DAGs.

The efficacy of the ECD approach is substantiated through experiments on both synthetic datasets and real-world EHRs. On synthetic datasets, ECD maintains high accuracy and stability across different noise levels, outperforming the conventional causal discovery methods. For the real-world EHR dataset, ECD reveals the intricate relationships between BMI and other predictive variables, aligning with the results of SEM and SHapley Additive exPlanations (SHAP) analyses. The RIS algorithm further quantifies the impact of perturbations on predictive variables, providing a basis for expression simplification and counterfactual reasoning.

The ECD framework represents an evolution and augmentation of existing causal discovery methods, providing an interpretable approach for analyzing variable relationships in complex systems, particularly in healthcare settings with EHR data. The proposed method demonstrates unique advantages in handling complex and structurally unknown data, promising to become a novel, interpretable method for complex data analysis, expanding the applicability of causal exploration in various domains.

The remainder of the paper is organized as follows. First, the related works are presented. Second, the methodology is presented. Third, the experiments and their results are presented, which are then followed by the discussions. Finally, the paper is concluded.

\section{Related Work}
\label{sec: related_work}

Evolutionary computation has been a cornerstone of optimization and search problems owing to its ability to handle complex, non-linear, and dynamic optimization challenges \cite{Coello2007, Coello_Wu2023, HeuristicLab2005, deap2012}. 

Coello et al. \cite{Coello2002EvolutionaryAF, Coello_Mndez2015} provided a foundational framework for evolutionary algorithms in solving multi-objective problems, highlighting the adaptability and efficiency of these methods in navigating the search space of potential solutions. In the realm of evolutionary optimization, Wu et al. \cite{Wei2023} demonstrated the effectiveness of simplified helper tasks, underscoring the method's utility in high-dimensional and resource-intensive problems.

Causal discovery methods aim to deduce potential causal relationships directly from data, typically represented as DAGs \cite{Hoyle1995, Pearl2000, Glymour2019}. These methods are crucial for understanding the complex interactions between health conditions and treatment outcomes in the absence of randomized controlled trials. The ECD method differs from conventional causal discovery by focusing on unveiling intrinsic mathematical relationships in data rather than directly seeking causal relationship graphs \cite{Glymour2019, DirectLiNGAM2011, NOTEARS2018}. These methods underscore the importance of structured graphical representations in elucidating causal pathways, a concept that aligns with the DAG-oriented approach of the ECD method.

As a tool for discovering mathematical expressions that best describe a dataset, symbolic regression intersects significantly with causal discovery \cite{Otsuka2016}. Orzechowski et al. \cite{Orzechowski2018} conducted a benchmark study on symbolic regression methods, elucidating their capabilities in identifying underlying data patterns. The ability of symbolic regression to generate interpretable models aligns with the objectives of ECD, emphasizing its importance in exploring and quantifying causal relationships.

The evolution of model discovery methods, exemplified by Gunaratne et al. \cite{Gunaratne2020}, highlights the innovative potential of integrating evolutionary computation and causal analysis. This convergence of evolutionary computation and causal analysis provides a rich backdrop for the ECD method, highlighting the innovative potential of combining these domains to advance our understanding of complex systems.

The use of EHR for medical and health research has become increasingly prevalent owing to the wealth of information it provides for understanding disease mechanisms and optimizing clinical decisions \cite{LaCava2023, Kotoku2020, Erion2022, Morin2021}. Zhao et al. \cite{Zhao2019} leveraged longitudinal EHR and genetic data to enhance cardiovascular event prediction, illustrating the potential of integrating diverse data sources for robust causal inference.

The ECD method is particularly adept at handling the complexities of EHR data, which often includes both numerical and categorical variables with potential missing values. This approach is consistent with the work of La Cava et al. \cite{LaCava2023}, who developed a flexible SR method for constructing interpretable clinical prediction models from EHR data.

The RIS algorithm proposed in this study systematically evaluates the effect of variable perturbations within a graph structure, providing a quantitative basis for expression simplification and refinement of key system mechanisms. This aligns with the work of Zheng et al. \cite{NOTEARS2018}, who developed a method for continuous optimization for structure learning in DAGs. The RIS algorithm extends the capabilities of ECD by enhancing the interpretability of system dynamics mechanisms and providing potential pathways for causal effect evaluation and counterfactual reasoning.

Recently, Bi et al. \cite{Bi2024} introduced the Genetic Programming (GP)-based evolutionary deep learning approach for data-efficient image classification and demonstrate the power of GP in learning interpretable models. These methods focus on prediction tasks, whereas the ECD method extends GP's advantages to causal discovery. ECD distinguishes itself by focusing on causal relationships and introducing customized post-processing techniques like RIS to enhance interpretability.

Zhang et al. \cite{Zhang2024} surveyed the integration of domain knowledge and machine learning techniques into GP for job shop scheduling, influencing the design of ECD. ECD incorporates domain-specific operators and machine-learning techniques to improve causal discovery efficacy. Moreover, it proposes new post-processing and visualization tools, such as RIS and expression tree, tailored for causal analysis. ECD adopts a unified design, synergizing evolutionary computation, causal inference, and machine learning principles throughout the modeling and optimization process.

Generally, ECD aims to utilize the foundational work in evolutionary computation, symbolic regression, EHR analysis, and causal discovery to advance the field of interpretable data analysis. The ECD framework offers a new approach for uncovering patterns and mechanisms among variables, particularly in complex and structurally unknown data. This work contributes to the theoretical understanding of causal relationships and has practical implications for clinical decision-making processes and policy development in healthcare.

\section{Methodology}

\subsection{Brief Concept}

The ECD analytical method encompasses the relationship between predictor and response variables. Predictor variables represent observations of specific aspects of a system, whereas response variables embody the comprehensive impact exerted by the predictor variables on the system. For instance, within the realm of health analytics and medicine, predictor variables can be construed as a set of health indicators, with the response variables corresponding to the diagnostic outcomes derived from these indicators. The ECD method distinctly emphasizes the perspective of the data analyst, prioritizing the predictive outcomes and delving into the underlying mechanisms governing the interaction between predictor and response variables.

The ECD method, utilizing GPSR for expression parsing, designs operators to abstract and connect variable relationships. As an advanced data-driven modeling strategy, the GPSR can autonomously reveal complex relationships within data, particularly suited for scenarios with numerous variables and undefined relationships. Without the need for a predefined model structure, the GPSR optimizes symbolic expressions by simulating natural selection and genetic mechanisms, offering a powerful analytical tool for scientific research. The technique has demonstrated extensive applicability in fields such as physics, environmental science, financial economics, and chemical and pharmaceutical design, with growing recognition in bioinformatics, genomics, EHR, and clinical medicine in recent years.

However, a limitation of the GPSR is the potential verbosity of its expressions, specifically when dealing with complex data relationships in EHR. Cumbersome expressions are not conducive to identifying key mechanisms or interpreting results. To address this, the ECD method proceeds with the RIS algorithm, which assists in analyzing relationships between variables by evaluating the impact of minor perturbations in predictor variables across statistical quartiles on various operator nodes and the response variable. Particularly for multi-level or effect-overlapping variable relationships, the RIS algorithm provides a concise quantitative basis for expression simplification, refining the articulation of key system mechanisms.

The ECD expression tree, equivalent to parsing expressions, visualizes the results of the RIS algorithm. The expression tree can be understood as a DAG connecting variables with operators, offering a differentiated depiction of the same unknown causal relationships compared to a DAG of conventional causal discovery, a topic further discussed in Section~\ref{sec: disc}. The expression tree enhances the interpretability of system dynamics mechanisms and extends potential pathways for causal effect evaluation and counterfactual reasoning.

\subsection{Prerequisites}

The ECD framework for evolutionary computation is predicated upon several foundational assumptions essential for its applicability to systems modeling. First, it presupposes a limited set of observations for state-describing variables, each bounded within a tolerable error margin. This suggests an acknowledgment of the inherent observational uncertainties in practical data collection. Second, variables within this framework are dichotomized into two distinct categories: a set of predictive variables and a singular response variable. The predictive variables could collaboratively affect the state of the system, with the caveat that there must be at least two such variables exhibiting a degree of interdependence, precluding any scenario of complete independence. Conversely, the response variable is intended to encapsulate the system's output or the resultant state as influenced by the interplay of all or a subset of the predictive variables. This bifurcation enables a clear delineation between inputs and outputs within the model, facilitating the identification of causal relationships and the derivation of predictive insights.

\subsection{GPSR}

ECD facilitates the design and implementation of evolutionary algorithms, including GP and genetic algorithms. The GPSR execution principles and workflow are described as follows.

\subsubsection{Population Initialization}
Let \(P(t)\) represent the population at generation \(t\), where \(P(0)\) is the initial population. This population consists of \(N\) individuals, i.e., \(P(t) = \{I_1, I_2, ..., I_N\}\), where each individual \(I_i\) is a program or an expression tree.

\subsubsection{Fitness Function}
The fitness function \(f: I \rightarrow \mathbb{R}\) maps an individual \(I\) to a real number, signifying its fitness or performance. The objective is to maximize (or minimize) this fitness value.

\subsubsection{Genetic Operations}
Genetic operations generate new individuals as follows:
\begin{itemize}
    \item \textbf{Selection} (\(\text{Sel}\)): This operation selects superior individuals based on fitness, forming a subpopulation \(P'(t)\) from \(P(t)\), i.e., \(\text{Sel}: P(t) \times \mathbb{R}^N \rightarrow P'(t)\).
    \item \textbf{Crossover} (\(\text{Cross}\)): It combines parts of two individuals to create offspring, leading to a new population \(P''(t)\), i.e., \(\text{Cross}: P'(t) \rightarrow P''(t)\).
    \item \textbf{Mutation} (\(\text{Mut}\)): This operation presents small random changes in an individual, producing the next generation population \(P(t+1)\), i.e., \(\text{Mut}: P''(t) \rightarrow P(t+1)\).
\end{itemize}

\subsubsection{Evolutionary Loop}
The evolutionary process can be expressed as an iterative loop in Eq.~ref{eq:fpt}:
\begin{equation}
\label{eq:fpt}
    P(t+1) = \text{Mut}\Big(\text{Cross}\big(\text{Sel}(P(t), f(P(t)))\big)\Big)
\end{equation}
where \(f(P(t))\) is the application of the fitness function to all individuals in the current population.

\subsubsection{Termination Condition}
The algorithm terminates when a condition \(C\) is satisfied, which could be reaching a maximum number of generations \(t_{\max}\), achieving a satisfactory solution, or the improvement of fitness falling below a threshold.

The entire GP process can be viewed as a dynamic system wherein the evolution of the population is influenced by genetic operations, continually converging toward an optimal solution.

\subsection{RIS}
\label{sec: RIS}
The RIS algorithm primarily serves two objectives. First, it furnishes a quantifiable reference for the simplification of GPSR expressions, enabling a systematic reduction of model complexity while retaining predictive accuracy. Second, it obtains a restricted counterfactual outcome associated with the predictive variables about the response variable, thereby elucidating the causal mechanisms underlying the observed data. This dual-faceted approach facilitates a deeper understanding of the model's structure and the causal relationships it encapsulates, enhancing the interpretability and robustness of the derived symbolic expressions.

The RIS algorithm systematically evaluates the effect of variable perturbations within a graph structure, representing a symbolic regression model. Let $G = (N, E)$ denote the graph, with $N$ as the set of nodes, representing variables and operations, and $E$ as the set of edges indicating the computational relationships between nodes. Given a set of initial values $V$ for the nodes and a set of perturbations $\Delta V$, the RIS algorithm quantifies the impact of these perturbations on the model's output.

For each variable $v \in N$ and its corresponding perturbation $\delta_v \in \Delta V$, the algorithm modifies the initial value of $v$ to obtain a perturbed value set $V'$. This perturbation is expressed in Eq.~\ref{eq:V}:
\begin{equation}
\label{eq:V}
    V' = V + \delta_v,
\end{equation}
where the perturbation is applied to the specific variable under consideration, and the rest of the values remain unchanged.

The core of the RIS algorithm lies in the computation of the graph's output for both the original and perturbed values, which is represented in Eq.~\ref{eq:impact}:
\begin{equation}
\label{eq:impact}
    \text{Impact}(v) = f(V'_{v}) - f(V),
\end{equation}
where $f(V)$ denotes the output of the graph $G$ with the original values and $f(V'_{v})$ represents the output with the perturbed values. The function $f$ encapsulates the computational logic of the graph, processing through the nodes according to the operations defined by the edges.

The iterative process of value propagation through the graph is central to the evaluation of $f(V)$. For each edge $e \in E$ connecting nodes $n_1$ and $n_2$ with an operation $\text{op}_e$, the value of $n_2$ is updated following the operation's rule of Eq.~\ref{eq:n}:
\begin{equation}
\label{eq:n}
    n_2' = \text{op}_e(n_1, n_2).
\end{equation}

This update process continues iteratively until the values of all nodes in the graph stabilize, reflecting the comprehensive impact of the perturbation. Thus, the RIS algorithm provides a mechanism to quantify the relative impact of each variable's perturbation on the overall model output, highlighting the variables that significantly affect the model’s behavior. The details of RIS are shown in Algorithm.~ \ref{alg: RIS}.

\begin{algorithm}
\caption{Relative Impact Stratification (RIS)}
\label{alg: RIS}
\begin{algorithmic}[1]
\REQUIRE $N, E, V$: Nodes, Edges, Baseline values
\ENSURE Impact of perturbations on output
\item[]
\STATE \textbf{Function} RIS($N, E, V, \Delta V$)
    \STATE $I \gets$ empty dictionary
    \FOR{$(v, \delta)$ in $\Delta V$}
        \STATE $V'_v \gets V_v + \delta$ \COMMENT{Perturbation value of $v$}
        \STATE $O \gets$ Eval($N, E, V$)
        \STATE $O' \gets$ Eval($N, E, V'_v$)
        \STATE $I[v] \gets O' - O$ \COMMENT{Compute impact of perturbation}
    \ENDFOR
    \RETURN $I$
\item[]
    
\STATE \textbf{Function} Eval($N, E, V$)
    \STATE $V_n \gets V$
    \WHILE{not stable $V_n$}
        \FOR{$(n_1, n_2)$ in $E$}
            \IF{defined $V_n[n_1]$ and $V_n[n_2]$}
                \STATE $V_n[n_2] \gets$ Calc($n_1$, $n_2$, $V_n$)
            \ENDIF
        \ENDFOR
    \ENDWHILE
    \RETURN $V_n$
\item[]

\STATE \textbf{Function} Calc($n_1$, $n_2$, $V_n$)
    \STATE $op \gets$ operation of edge $(n_1, n_2)$
    \RETURN $op(V_n[n_1], V_n[n_2])$ \COMMENT{Apply operation}

\end{algorithmic}
\end{algorithm}

\section{Experiment}

\subsection{Experiment Design}
To assess the efficacy and interpretability of the ECD method, this study conducted two experiments: one based on synthetic datasets and the other on real-world EHR datasets.

The synthetic data comprised samples with predefined causal relationships among variables. By altering parameters such as sample size, complexity of causal relationships, and noise levels, we evaluated the ECD method's capacity to reconstruct known causal structures across diverse scenarios. Classical causal discovery algorithms fall into four main categories: constraint-, score-, functional causal modeling-, and structural causal model-based algorithms. From these categories, we selected one representative method each—Peter-Clark algorithm (PC), Greedy Equivalence Search (GES), Direct Linear Non-Gaussian Acyclic Model (Direct LiNGAM), and Non-combinatorial Optimization via Trace Exponential and Augmented lagRangian for Structure learning (NOTEARS) algorithm—for comparative analysis.

In the experiment utilizing real-world EHR data, the study employed the public dataset, `Indicators of Heart Disease (CDC2022)," obtained from the Centers for Disease Control and Prevention (CDC) and constituted a significant portion of the Behavioral Risk Factor Surveillance System (BRFSS)\footnote{The Behavioral Risk Factor Surveillance System (BRFSS) is the nation’s premier system of health-related telephone surveys that collects state data about U.S. residents regarding their health-related risk behaviors, chronic health conditions, and use of preventive services. The official website of BRFSS is located at \url{https://www.cdc.gov/brfss/}.}, which conducts annual telephone surveys to gather data on the health status of over 400,000 U.S. residents. We extracted a subset of this data and applied the ECD method to uncover latent causal relationships and interaction patterns among health-related variables.


Experiments were conducted in a Python environment (version 3.8 or higher), employing the `pgmpy' library for PC and GES algorithms, `lingam' for Direct LiNGAM, `CausalNex' for NOTEARS.

\subsection{Experiment on Synthetic Dataset}
\subsubsection{Dataset Generation}
\begin{figure}[h]\centering 
\includegraphics[width=0.6\linewidth]{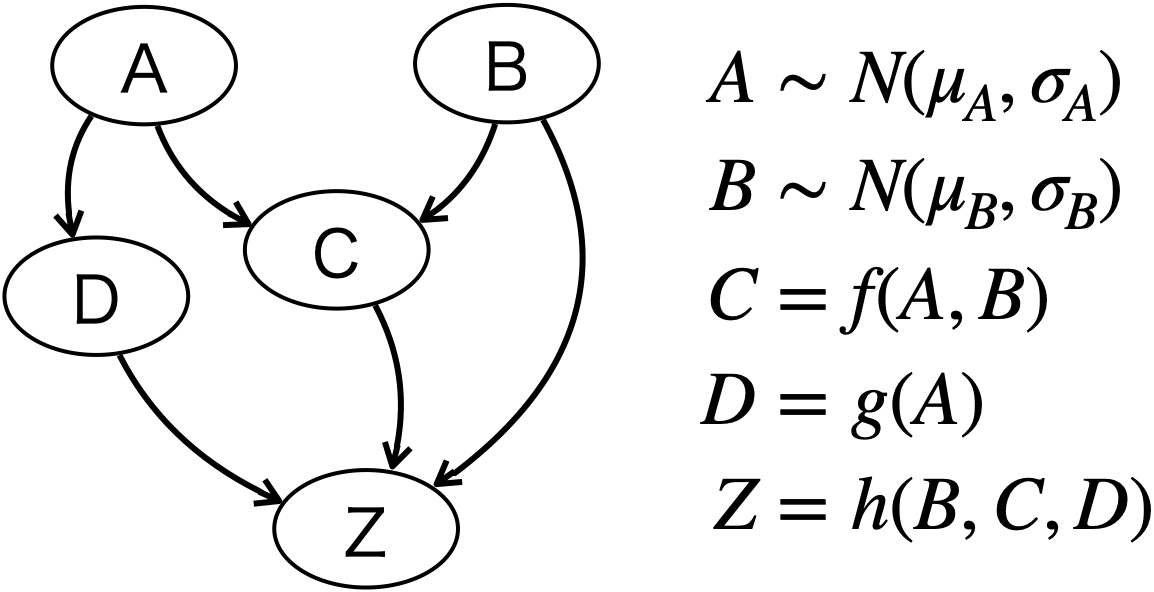}
\caption{Synthetic dataset with predefined causal relationships among variables, i.e. unknown ground truth. As a simple example for the methodological test, predictive variables $A$ and $B$ follow normal distributions $A \sim \mathcal{N}(1,2)$ and $B \sim \mathcal{N}(2,1)$, respectively. Derived variables are $C = A + B$, $D = 2A + 3$, and response variable $Z = B + \frac{C}{D}$. Sample size $n = 500$, with noise levels at 0\%, 2\%, and 5\%.}
\label{fig: syn_1}
\end{figure}

The study generates a synthetic dataset with four predictive variables ($A, B, C, D$) and a response variable ($Z$), using a sample size $n$ and a random seed for reproducibility. Variables $A$ and $B$ are normally distributed. $C$ and $D$ are linearly derived from $A$ and $B$, respectively, whereas $Z$ presented non-linearity. The causal structure shown in Fig.~\ref{fig: syn_1} includes basic causal units like Chains (a direct line of influence from one variable to the next, e.g.:$A\to D\to Z$), Colliders (two or more variables influencing a common effect, e.g.: $A\to C\gets B$), and Forks (a common cause affecting two or more variables, e.g.: $D\gets A\to C$), with additional cross-layer interactions (e.g.: $B\to E$). To simulate real-world noise, a function applied Gaussian noise to predictive variables based on a set noise percentage and the absolute value of data.

\subsubsection{Result and Interpretation}

\begin{figure}[h]\centering 
\includegraphics[width=1.0\linewidth]{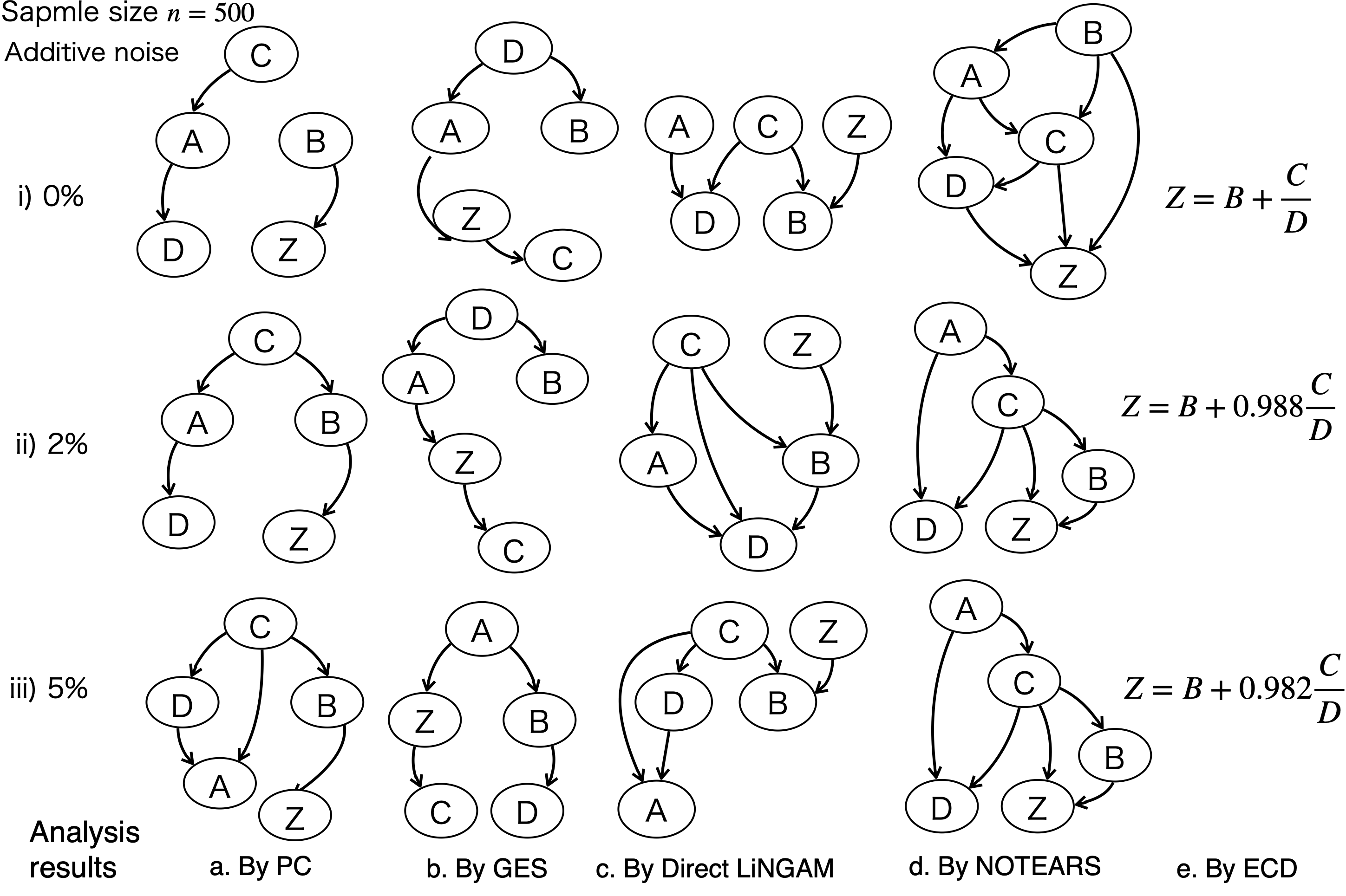}
\caption{Experimental results of selected major causal discovery methods and ECD on the synthetic dataset, as shown in Fig.~\ref{fig: syn_1}.}
\label{fig: syn_1_result_500}
\end{figure}

Fig.~\ref{fig: syn_1_result_500} presents the inferential outcomes of four prevalent causal discovery algorithms (PC, GES, Direct LiNGAM, and NOTEARS), as well as ECD utilizing GPSR on the synthetic dataset. The algorithms are anticipated to accurately unveil the predetermined causal relationships among the variables, aligning closely with the ideal causal structure depicted in Fig.~\ref{fig: syn_1}, particularly in the precise identification of key causal directions. Under various levels of noise interference, the DAG structures inferred by the NOTEARS and PC algorithms demonstrated notable precision and stability in delineating $Z$ as the response variable and in revealing the principal relationships among variables. Nevertheless, the NOTEARS algorithm exhibited uncertainties in discerning the causal directions between $C$ and $D$; similarly, the PC algorithm did not accurately identify the causal directions between $A$ and $C$, as well as $B$ and $C$. The GES algorithm provided consistent DAG results under 0\% and 2\% noise conditions, with slight variations emerging at the 5\% noise level. Direct LiNGAM's performance on the synthetic dataset was inferior, recognizing some variable relations but displaying discernible discrepancies from the true causal structure.

Notably, the causal structures inferred by different algorithms can vary significantly because of the distinct assumptions, statistical tests, data characteristics, and sample sizes. In practical applications, these variations necessitate thorough analysis and judicious interpretation. Hence, the results depicted in Fig.~\ref{fig: syn_1_result_500} cannot be utilized to categorically evaluate the superiority or inferiority of the various causal discovery algorithms, reflecting the inherent challenges in evaluating causal discovery methods in applied settings. For instance, the Direct LiNGAM has been successfully applied in the research\cite{Kotoku2020}, concerning large-scale health check datasets, seamlessly integrating with statistical analytical methods.

Nonetheless, the results from Fig.~\ref{fig: syn_1_result_500} indicate that the ECD method maintains high-resolution accuracy and stability across all examined noise levels. In a noise-free environment, ECD was able to infer the precise causal structure from the data, and it continued to sustain a near-ideal solution when noise increased to 5\%. These findings suggest that the ECD approach based on GPSR, possesses potential advantages in the identification of causal relationships, particularly when dealing with smaller sample sizes and a specific group of variables.

\subsubsection{ECD Training Procedures}

\begin{figure}[h]\centering 
\includegraphics[width=1.0\linewidth]{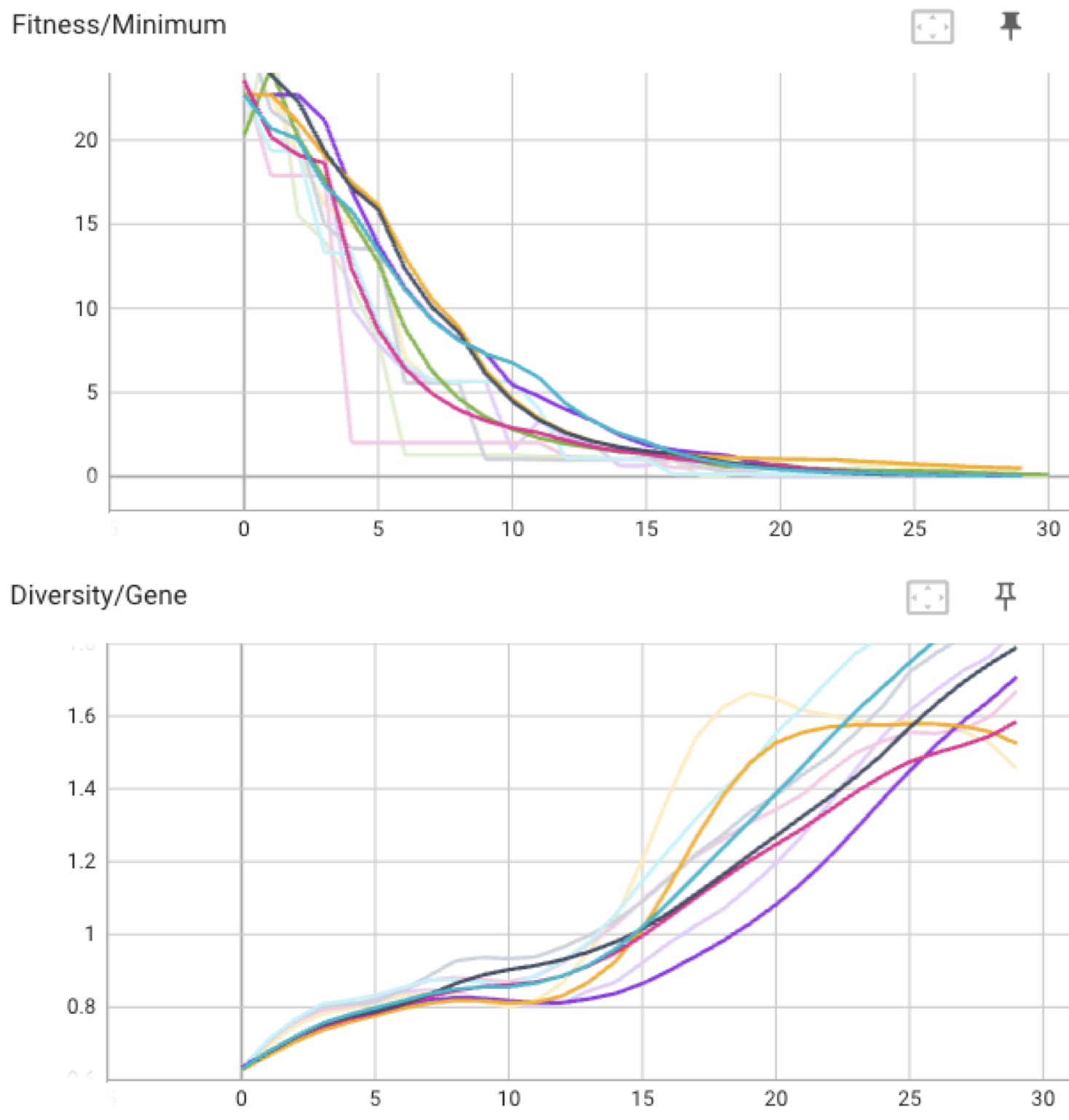}
\caption{Evaluation of ECD training procedures on the experimental synthetic dataset. The figure illustrates the evolutionary trajectories of minimal fitness and genetic diversity, depicted as curves of distinct colors, across ten experimental runs. Each experiment is conducted with a set duration of 30 generations.}
\label{fig: GPSR_train_1}
\end{figure}

Fig.~\ref{fig: GPSR_train_1} delineates the evolutionary trajectories of fitness and genetic diversity across generations during multiple runs of ECD utilizing GPSR. The fitness graph illustrates a consistent downward trend in the minimal fitness values within the population, indicating an optimization of solutions converging towards an ideal model as the evolutionary process unfolds. The convergence of fitness values suggests that the algorithm is steadily progressing towards a solution that minimizes error.

In contrast, the diversity graph charts an upward trajectory in genetic diversity, indicating  the algorithm's initial robust exploration of the solution space. A sustained increase in diversity could be indicative of the population's strategy to circumvent local optima and enhance global search efforts.

The intersection point, observed at approximately the 10th generation, marks a pivotal phase where improvements in fitness coincide with an increase in diversity. This may reflect the algorithm's exploration of new regions within the solution space, aiming for the global optimum. Collectively, these outcomes suggest that while the GPSR exhibits potential in iteratively identifying optimal solutions, the rising diversity within the population also denotes the algorithm's ongoing search-exploit balance. This underlines the necessity of continuous monitoring of these trends to ensure the development of a robust final model.

\subsection{Experiment on Real Dataset}
\subsubsection{Variable Set Selection and Preliminary SEM Analysis}

Here, we aim to validate the efficacy of the ECD approach using an actual EHR dataset, rather than conducting an in-depth investigation of specific medical or health-related issues. We focus on males aged 60 to 69 from the CDC2022 dataset, which includes 26143 entities, with a selected variable set of ”BMI,” ”GeneralHealth,” ”SmokerStatus,” “AlcoholDrinkers,” and “SleepHours” for analysis. The descriptions of these variables are enumerated below.

\begin{itemize}
    \item \textbf{AlcoholDrinkers:} coded as `Yes' for 1, and `No' for 2;
    \item \textbf{GeneralHealth:} rated on a scale with `Poor' as 1, `Fair' as 2, `Good' as 3, `Very good' as 4, and `Excellent' as 5;
    \item \textbf{SmokerStatus:} categorized into `Current smoker-now smokes every day' as 1, `Current smoker-now smokes some days' as 2, `Former smoker' as 3, and `Never smoked' as 4;
    \item \textbf{SleepHours:} an integer variable representing the number of hours of sleep;
    \item \textbf{BMI:} a continuous numerical variable representing the Body Mass Index.
\end{itemize}

We aim to examine the interrelations between these variables and BMI, and based on the outcomes of GPSR, to further assess the response sensitivity of BMI to controlled variations in specific predictive variables through the RIS algorithm.

The careful selection of the variable set includes various patterns of correlation for experimental purposes, which are shown in the preliminary SEM analysis results in Table.~\ref{tab: SEM_1}. SEM is a multivariate statistical technique that reveals the relationships between multiple measured variables and latent constructs. In our experiments, SEM was employed to test the statistical significance of the relationships between BMI and other predictive variables. In Table.~\ref{tab: SEM_1}, these indicators collectively reflect the effectiveness of the model and a high level of consistency with the data.

\begin{table*}
\centering
\caption{Analysis of SEM model-data-fit for response variable BMI and other predictive variables}
    \begin{tabular}{lclrrrrl}
   \toprule
\textbf{Resp. Var} &\textbf{op}& \textbf{Pred. Var} & \textbf{Estimate} & \textbf{Std. Err} &\textbf{z-value}&\textbf{p-value} &\textbf{Correlation Result} \\
\hline
BMI &$\sim$& GeneralHealth & -1.2236 & 0.0332 & -36.7908&$<0.0001$ &strong negative\\
BMI &$\sim$& SmokerStatus & 0.7983 & 0.0368 & 21.7217 &$<0.0001$ & strong positive\\
BMI &$\sim$& AlcoholDrinkers & 0.3334 & 0.0689 & 4.8377 &$<0.0001$& Relative weak positive\\
BMI &$\sim$& SleepHours & -0.0359 & 0.0234 & -1.5375&$0.1242$ &no significant \\
\hline
\multicolumn{2}{l}{Optimization result} &\multicolumn{5}{l}{Method: SLSQP; Objective value: 0.000; Number of iterations: 28} &\\ 
\multicolumn{2}{l}{Goodness-of-Fit Measures}&\multicolumn{5}{l}{$\chi^{2}$statistic:  0.0082; CFI:1.0026; GFI: 0.9999; AIC: 9.99; BIC: 50.86}& Strong model-data-fit\\ 
\bottomrule
\end{tabular}
  \label{tab: SEM_1}
\end{table*}

\subsubsection{Result and Interpretation}

\begin{figure*}[h]\centering 
\includegraphics[width=0.9\linewidth]{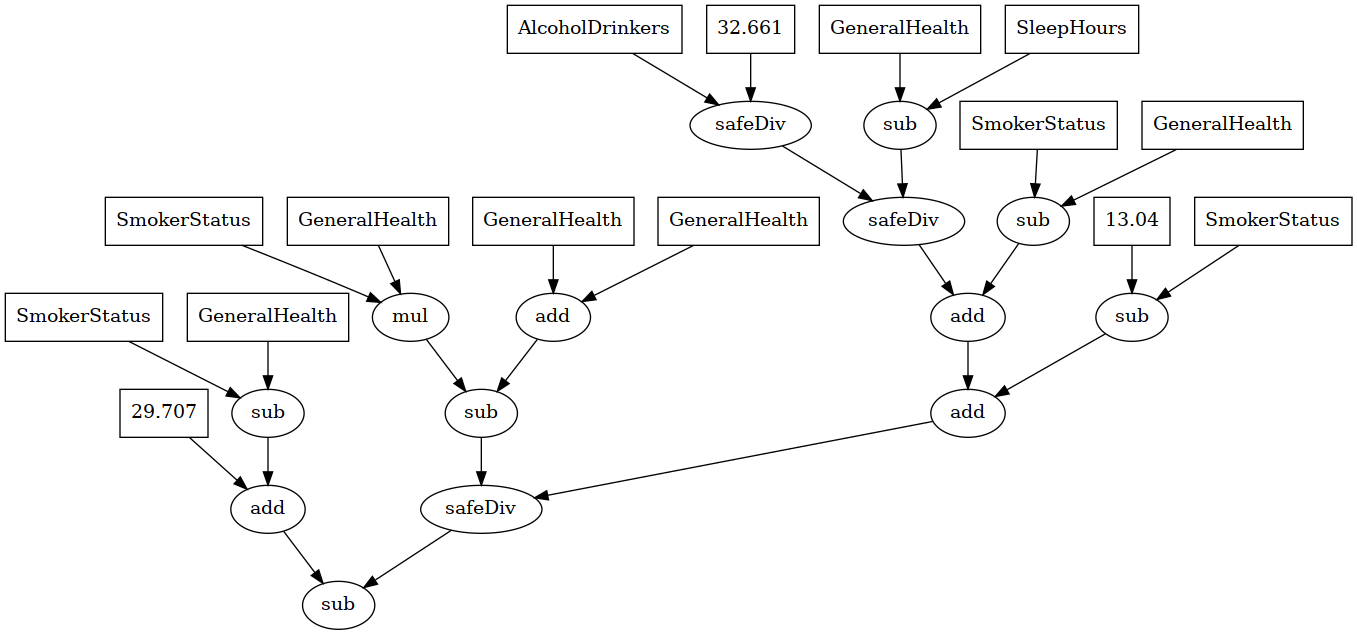}
\caption{Expression tree of ECD analysis of the experimental EHR dataset. Box nodes represent predictive variables situated at the topmost leaf nodes of the expression tree, whereas ellipse nodes denote symbolic regression operators constituting the internal nodes of the tree.}
\label{fig: GPSR_expr}
\end{figure*}

Considering the outcomes and interpretability of preliminary SEM analysis, this study predominantly adopts linear hypotheses for setting the operators in symbolic regression. Unlike the straightforward results on synthetic datasets, the complexity of real EHR data significantly increases, leading to more complex symbolic regression expressions.

Symbolic regression expressions function as a unique form of DAG, as exemplified by the expression trees depicted in Fig.~\ref{fig: GPSR_expr}, containing operators. These differ from conventional causal discovery DAGs, reflecting distinct causal assumptions and modeling approaches. In conventional causal DAGs, nodes represent variables and edges denote direct causal relationships among these variables. Such DAGs assume that variable relationships can be depicted with a DAG and utilize conditional independence tests or score-based methods for learning the DAG structure. The conventional causal DAGs aim to uncover causal dependencies among variables without specifying the functional form of causal relationships explicitly.

In contrast, the symbolic regression expression trees in ECD, also known as operator-inclusive DAGs or simply expression trees, feature internal nodes representing mathematical operations or functions, with leaf nodes denoting variables or constants. These expression trees, used for predicting or fitting target variables, aim to reveal the mathematical relationships among variables and explicitly specify the functional form of these relationships.

Fig.~\ref{fig: GPSR_expr} shows an expression tree that represents the response variable BMI using selected predictive variables. This tree, derived from preset prediction accuracy and operators via GP, represents the optimal function expression found within the potential solution space. It exhibits the following characteristics: (1) a top-down data flow structure without reverse flow or loops; (2) variables or constants only appear at leaf nodes, with mathematical operations and functions at internal nodes; (3) the tree's hierarchical structure, which equates to the mathematical operation's precedence order, as reflected in the analytical expression.

\subsubsection{Application of RIS}

In Section \ref{sec: RIS}, we discussed two primary applications of RIS: simplification of expression trees and counterfactual reasoning with predictive variables. The fundamental principle of RIS involves introducing perturbations to predictive variables and observing the consequent changes in the internal nodes of the expression tree. Because predictive variables occupy the topmost leaf nodes, their perturbations propagate downstream, amplifying or attenuating until manifesting at the final nodes, either directly affecting the response variable BMI or dissipating within the internal nodes.

RIS unveils the details of how local perturbations can induce global changes, enriching the analysis of static data with dynamic insights and aiding in partially uncovering the system dynamics. RIS focuses on evaluating relative impact effects, necessitating the establishment of a baseline for comparison. The methodologies for setting this baseline include: (1) using domain knowledge and specific conditions to assign certain values to predictive variables for RIS computation; (2) employing statistical quartiles as the baseline to perturb predictive variables, the latter of which facilitates causal discovery without prior knowledge.

\begin{figure}[h]\centering 
\includegraphics[width=1.0\linewidth]{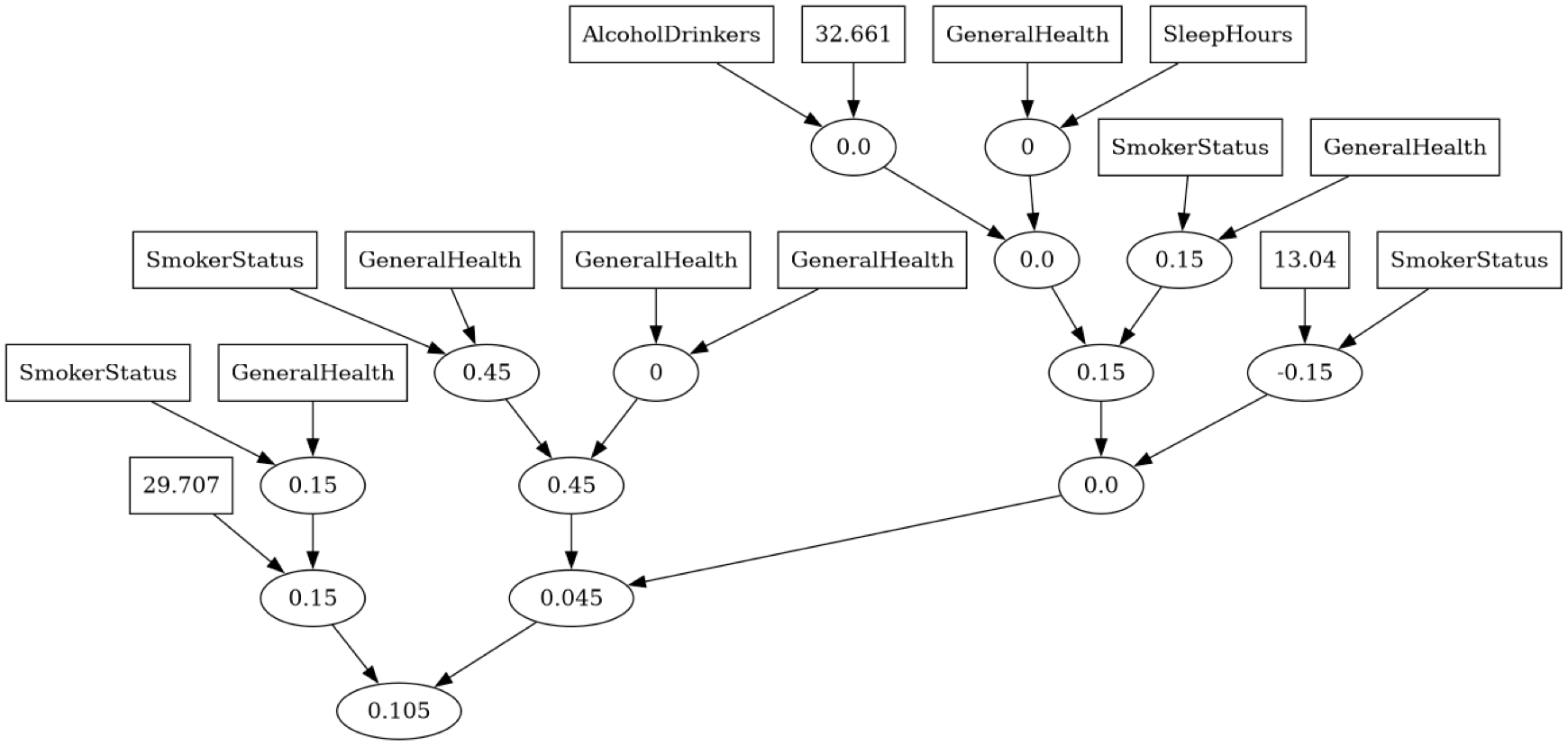}
\caption{Exploratory analysis conducted via RIS within the ECD method. A perturbation analysis of 5\% positive as an example was performed on the predictive variable `SmokerStatus' with the 1st quartile of predictive variables.}
\label{fig: RIS_result}
\end{figure}

In our experiments, we adopt the latter approach to elucidate RIS. For instance, Fig.\ref{fig: RIS_result} illustrates the effects of a 5\% perturbation at the first quartile on the SmokeStatus variable, signifying a minor improvement in SmokeStatus, which is expected to result in a 0.105 increase in the BMI. Furthermore, an analysis of the expression tree's internal nodes reveals the predominant effect of the interaction between SmokerStatus and GeneralHealth on the BMI.

Employing this method, we can derive the RIS analysis of perturbation effects for all predictive variables across quartiles. As demonstrated in Table.~\ref{tab: RIS_result_1}, these results align with the SEM fit findings from Table.\ref{tab: SEM_1}; however, they provide additional details. For example, improvements in GeneralHealth at the third quartile could lead to the most significant relative reduction in the BMI. Minor enhancements in SmokerStatus are effective for individuals of moderate health levels. The slight changes in AlcoholDrinker and SleepHours variables have no significant impact on the BMI across all quartiles.

Given the heterogeneity and complexity of EHR data, the quartile-based RIS analysis serves as a statistical reference, with its reliability pending clinical or further experimental verification. The RIS outcomes offer a quantified basis for directional choices in subsequent validations.

\begin{table}
\centering
\caption{Results of RIS analysis of predictive variables on the response variable BMI}
\begin{tabular}{lrrr}
\toprule
\textbf{Predictive Variable} &\multicolumn{3}{c}{\textbf{Impact on BMI at Quartiles}} \\
(5\% positive perturbation)   &\textbf{1st }& \textbf{2nd} & \textbf{3rd} \\
\hline
 GeneralHealth & - 0.170 & - 0.189 & - 0.266\\
 SmokerStatus & + 0.105 & + 0.140 & + 0.111 \\
 AlcoholDrinkers &$\pm$ 0.0 &$\pm$ 0.0 &$\pm$ 0.0  \\
SleepHours & $\pm$ 0.0 & $\pm$ 0.0 &$\pm$ 0.0  \\

\hline
\textbf{BMI Calculated Baseline$^*$}&29.4&30.1&28.8\\

\bottomrule
\multicolumn{4}{l}{$^*$Calculated by the expression tree shown in Fig.~\ref{fig: GPSR_expr}.}\\
\end{tabular}
  \label{tab: RIS_result_1}
\end{table}

\section{Discussion}
\label{sec: disc}

\subsection{Operator Setting of ECD}
In our preliminary experiments, we utilized only basic arithmetic operators to validate principles. GPSR extends the ECD method by proceeding with complex mathematical functions, logical operations, and conditional operators such as if-then-else. However, the introduction of complex  operators can complicate the expression model. Therefore, it is essential to select an optimal operator set based on problem characteristics to maintain an efficient and reasonable search space.

In healthcare, where interpretability is crucial, simpler or linear operators often enhance understandability.

Crucially, ECD is analyzed with domain knowledge operators, which align with specific domain priors to add semantic depth to the expression tree. Examples include medical risk assessment operators that calculate disease risk scores using inputs like age, BMI, and family history; time series modeling operators for continuous physiological data such as heart rate and blood pressure; survival analysis operators for modeling patient survival data.

These customized operators enable GP to automatically discover clinical models beyond simple mathematical combinations, enhancing causal discovery in healthcare. This approach necessitates close collaboration with domain experts to ensure the practicality and effectiveness of the algorithm.

\subsection{Hyperparameter Setting of ECD}

The configuration of hyper-parameters for the ECD method is described as follows:

For synthetic datasets, the settings include population size of 50,000, crossover probability of 0.5, mutation probability of 0.1, and a number of generations set to 30.

For real datasets, adjustments are made as follows: the population size is increased to 100,000; the crossover probability is adjusted to 0.6; the mutation probability is set to 0.2; the number of generations remains at 30. These adjustments aim to enhance the search capabilities and adaptability of the algorithm.

Directions for parameter adjustment considered in experiments include:

\begin{itemize}
\item{Population Size: Increasing the population size helps provide a broader range of initial solutions, enhancing the algorithm’s ability to explore the solution space; it is particularly beneficial in noisy environments to maintain diversity.}
\item{Crossover Probability: Elevating the crossover probability promotes diversity within the population, aiding the algorithm in exploring an extensive solution space and preventing premature convergence on local optima.}
\item{Mutation Probability: Raising the mutation probability increases the diversity of the population, enabling the algorithm to discover more unexplored areas and enhancing its capacity to adapt to noisy conditions.}
\item{Number of Generations: Extending the number of generations allows the algorithm more time to adapt to and overcome the impacts of noise, seeking optimal solutions.}
\end{itemize}

Experiments demonstrate that increased parameter values do not necessarily yield positive effects and may sometimes lead to prolonged computation times or imbalances in processing efficiency, particularly as an increase in the number of generations can rapidly escalate the complexity of the expression trees, affecting interpretability. Consequently, there is no one-size-fits-all standard for parameter adjustment. The best practice is to determine the most suitable parameter settings for the current problem and data characteristics through iterative experimentation. One may also consider employing automated parameter search techniques, such as grid search or Bayesian optimization, to assist in identifying the optimal combination of parameters. By adjusting parameters and focusing on the diversity and exploratory behavior of the algorithm, its performance in complex environments can be effectively enhanced.

\subsection{Extending RIS in Counterfactual Reasoning}

The RIS algorithm applies perturbations to assess impacts under stable baseline conditions, inspired by Judea Pearl's do-calculus and counterfactual reasoning. Our experiments utilize a 5\% perturbation, chosen to avoid significant deviations from other variables while maintaining stability. This parameter reflects typical fluctuations found in health indicators such as heart rate and blood pressure, which often exhibit broad normal ranges. For variables with narrower fluctuation margins, like arterial oxygen saturation (SaO2), which typically ranges from 97\%$\sim$100\% in healthy conditions, caution is exercised to prevent unrealistic perturbations.

Additionally, RIS enables the design of initial inputs tailored to specific scenarios for counterfactual reasoning, which depends on a thorough understanding of underlying causal mechanisms—a process facilitated by the ECD method with RIS.

Building upon the experimental results presented in the previous sections, we can design a specific scenario to validate the application of the RIS algorithm in counterfactual reasoning. Let us consider a highly healthy initial state, where, based on the variable descriptions outlined earlier, we set the AlcoholDrinkers value to two (non-drinker), GeneralHealth value to five (excellent), SmokerStatus value to four (never smoked), and SleepHours value to eight (normal sleep duration). The counterfactual reasoning question is as follows: if an individual suddenly starts smoking, i.e., the `SmokerStatus' value changes to one (current smoker - now smokes every day), how might their BMI be affected? 

\begin{figure}[h]\centering 
\includegraphics[width=1.0\linewidth]{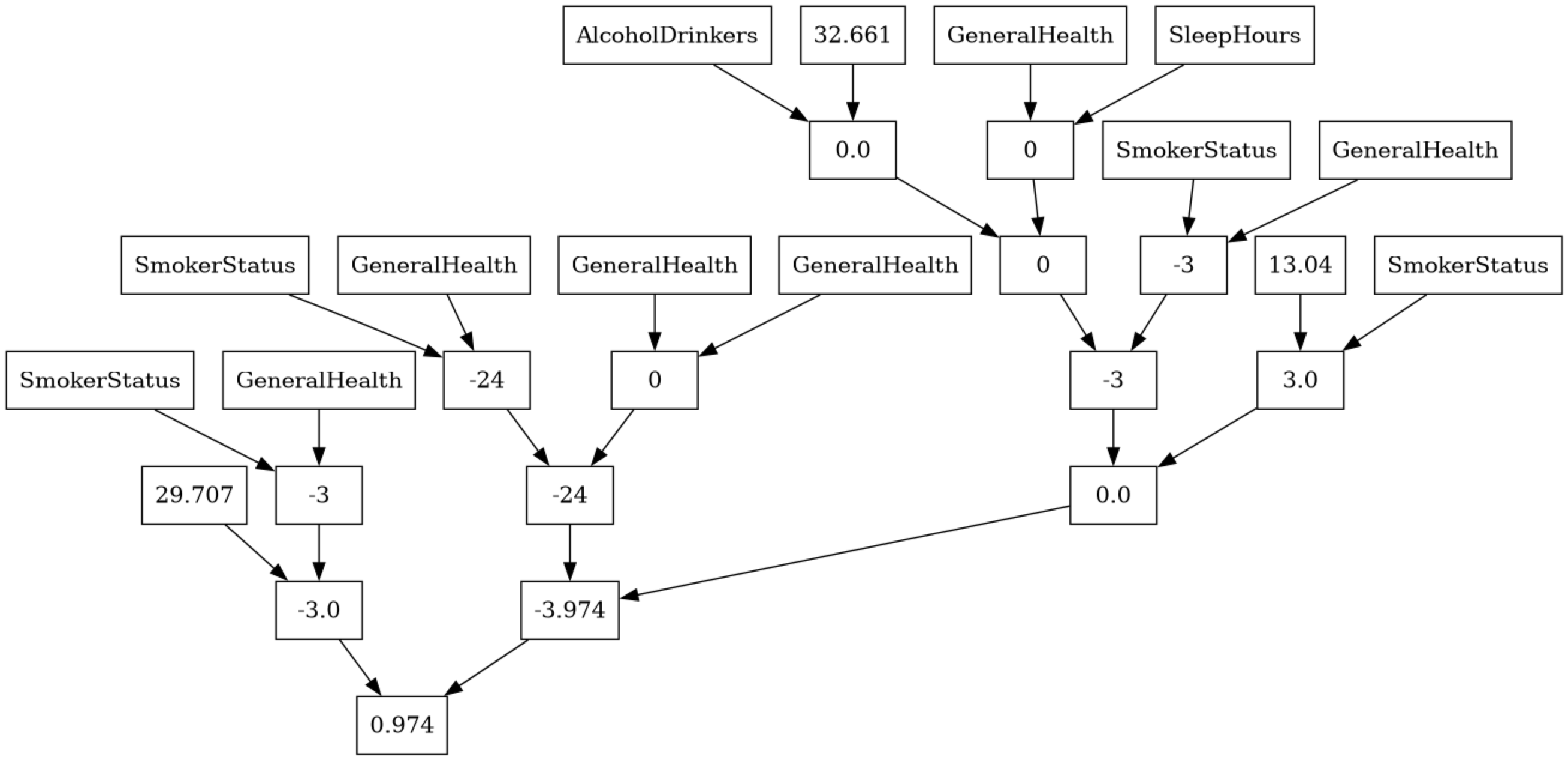}
\caption{Counterfactual reasoning scenario using the RIS algorithm. By adjusting the input value of SmokerStatus and observing the changes in the parent nodes and the response variable, the RIS algorithm quantifies the potential impact of a sudden change in lifestyle habits on the BMI.}
\label{fig: RIS_counterF_1}
\end{figure}

Thus, by inputting these hypothetical conditions into the ECD analytical expression, we obtain a BMI calculated baseline of 27.46 for this highly healthy state, which, according to the statistical results from the original dataset, is a relatively optimal level. Subsequently, we adjust the input value of `SmokerStatus' to one, and the RIS algorithm yields the results presented in Fig.~\ref{fig: RIS_counterF_1}, indicating that the BMI will increase by 0.974. Concurrently, we observe significant changes in both parent nodes of the final inner node. Specifically, owing to the change in SmokerStatus, its interaction with GeneralHealth becomes pronounced than before, thereby triggering a change in the response variable, the BMI. Therefore, we can infer that a sudden change in SmokerStatus may increase the BMI, which could further affect GeneralHealth. 

Notably, such hypotheses are difficult to directly verify in reality owing to ethical considerations. However, these results still hold value for scientific research, as they can help assess the potential changes that may arise from abrupt alterations in lifestyle habits. In such scenarios, the ECD method can play a role in predicting the possible trends of change in the response variable and identifying the most likely associated variables and pathways of influence.

Notably, counterfactual reasoning, guided by RIS, depends on ensuring the rationality of input and perturbation designs and requires consideration of variable thresholds owing to their complex interrelationships in health data. Analytical methods like breakpoint regression may help in identifying threshold ranges, critical for elucidating causal relationships in health analytics.

\subsection{Relationship of SHAP analysis and RIS}  

\begin{figure}[h]\centering 
\includegraphics[width=1.0\linewidth]{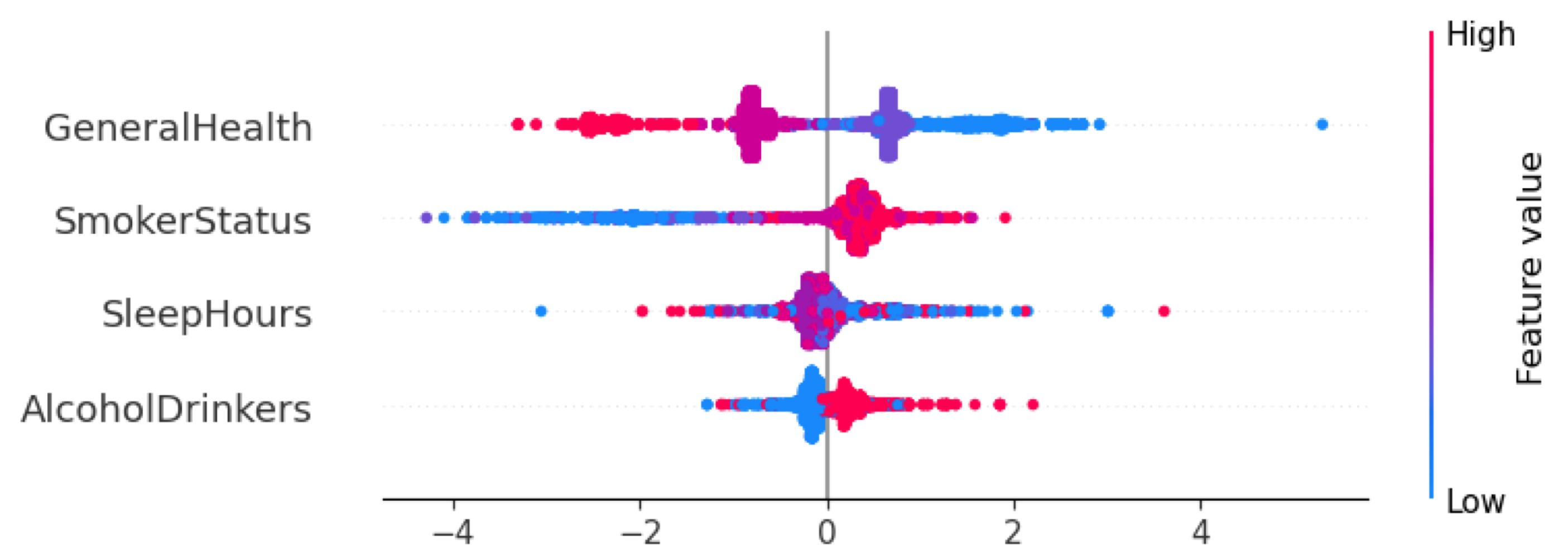}
\caption{SHAP analysis result of predictive variables on response variable BMI according to the experimental dataset.}
\label{fig: SHAP_1}
\end{figure}

SHAP values constitute a robust quantitative framework for assessing the impact of predictive variables on response variables, elucidating their marginal contributions through additive attribution values. Fig.~\ref{fig: SHAP_1} features a SHAP summary plot that synthesizes the effects of predictive variables on response variables, with each plot point corresponding to an observational instance and signifying its impact in terms of magnitude and direction.

Quantified by SHAP values on the horizontal axis, the marginal contributions of predictive variables are directly associated with their effect on the predictive accuracy of response variables, reflected in the polarity and magnitude of the SHAP values. The gradational color scheme from blue to red visually distinguishes variable values, illustrating their differential impact. The aggregation of SHAP values across the predictive variable axes suggests a concentrated distribution of impacts, with the bidirectional spread around zero indicating variable influences on prediction outcomes.

The analysis reveals a significant correlation between the GeneralHealth variable and the prediction outcomes, with higher SHAP values in specific instances suggesting an enhancement of predictive accuracy. Conversely, SmokerStatus appears to exert minimal influence, as indicated by SHAP values predominantly near zero, with occasional outliers signifying potential impacts under certain conditions. SleepHours demonstrates a dualistic effect, and `AlcoholDrinkers' indicates a generally marginal influence on prediction outcomes.

The alignment of the RIS algorithm with SHAP findings substantiates its analytical precision, particularly given its ability to discern finer details under variable conditions. The integration of RIS with SEM and SHAP analysis provides a comprehensive understanding of the predictive determinants in model outcomes, thereby offering a nuanced perspective in the interpretation of complex predictive decision-making processes.

\section{Conclusion}

This study proposes ECD, an approach for causal discovery that tailors response variables, predictor variables, and operators to research datasets. ECD utilizes GP for variable relationship parsing and proceeds with the RIS algorithm to enable expression simplification and enhance the interpretability of variable relationships.

Experiments on both synthetic and real-world EHR datasets demonstrate the efficacy of ECD in uncovering patterns and mechanisms among variables. On synthetic datasets, ECD maintains high accuracy and stability across different noise levels, outperforming conventional causal discovery methods. For the real-world EHR dataset, ECD reveals the intricate relationships between BMI and other predictive variables, aligning with the results of SEM and SHAP analyses. The RIS algorithm further quantifies the impact of perturbations on predictive variables, providing a basis for expression simplification and counterfactual reasoning.

The ECD expression tree, equivalent to parsing expressions, visualizes the results of the RIS algorithm. These graphs offer a differentiated depiction of unknown causal relationships compared to conventional causal discovery DAGs. The ECD method represents an evolution and augmentation of existing causal discovery methods, offering an interpretable approach for analyzing variable relationships in complex systems, particularly in healthcare settings with EHR data.

Although the ECD method demonstrates promising results, there are certain limitations to be addressed in future work. The current study focuses on a specific subset of variables from the EHR dataset, and expanding the analysis to a broader range of variables and datasets would further validate the generalizability of the ECD approach. Additionally, the operator settings in the experiments were primarily limited to basic arithmetic operators, and utilizing more complex mathematical functions, logical operations, and domain-specific operators could enhance the expressiveness of the models and uncover more intricate variable relationships.

Future research directions include collaborating with domain experts to embed domain-specific operators, investigating the integration of the ECD with other causal discovery and machine learning techniques, exploring the potential of the ECD in longitudinal studies, and applying the ECD to diverse datasets across different domains. By addressing these limitations and pursuing the outlined research directions, the ECD method can be further refined and extended to provide a powerful and interpretable framework for causal discovery and analysis in various complex systems.

\section*{Acknowledgments}
The work was supported in part by the 2022-2024 Masaru Ibuka Foundation Research Project on Oriental Medicine, 2020-2025 JSPS A3 Foresight Program (Grant No. JPJSA3F20200001), 2022–2024 Japan National Initiative Promotion Grant for Digital Rural City, 2023 and 2024 Waseda University Grants for Special Research Projects (Nos. 2023C-216 and 2024C-223), 2023-2024 Waseda University Advanced Research Center Project for Regional Cooperation Support, and 2023-2024 Japan Association for the Advancement of Medical Equipment (JAAME) Grant.

\bibliographystyle{IEEEtran}
\bibliography{references} 

\end{document}